\documentclass[10pt,twocolumn,letterpaper]{article}

\usepackage{iccv}              % To produce the CAMERA-READY version
\definecolor{iccvblue}{rgb}{0.21,0.49,0.74}
\usepackage[pagebackref,breaklinks,colorlinks,allcolors=iccvblue]{hyperref}

\usepackage{algorithm} 
\usepackage{algpseudocode} 
\usepackage{amsmath}
\usepackage{amssymb}
\usepackage{bm}
\usepackage{colortbl}
\usepackage{enumitem}
\usepackage{graphicx}
\usepackage{makecell}
\usepackage{multirow}
\usepackage{nccmath}
\usepackage{soul}
\usepackage{tabularx}
\usepackage{wrapfig}
\usepackage{xcolor}
\usepackage{xspace}

\newcommand{\first}[1]{\textcolor{blue}{\textbf{#1}}}
\newcommand{\second}[1]{\textcolor{black}{\textbf{#1}}}

\def\paperID{6666} % *** Enter the Paper ID here
\def\confName{ICCV}
\def\confYear{2025}

\title{TextSR: Diffusion Super-Resolution with Multilingual OCR Guidance}

\author{Keren Ye\quad Ignacio Garcia Dorado\quad Michalis Raptis\quad Mauricio Delbracio \\
Irene Zhu\quad Peyman Milanfar\quad Hossein Talebi
\\[.5em]
Google
}

\begin{document}
\maketitle
\begin{abstract}
\end{abstract}
\vspace{-8mm}

While recent advancements in Image Super-Resolution (SR) using diffusion models have shown promise in improving overall image quality, their application to scene text images has revealed limitations. These models often struggle with accurate text region localization and fail to effectively model image and multilingual character-to-shape priors. This leads to inconsistencies, the generation of hallucinated textures, and a decrease in the perceived quality of the super-resolved text.

To address these issues, we introduce TextSR, a multimodal diffusion model specifically designed for Multilingual Scene Text Image Super-Resolution. TextSR leverages a text detector to pinpoint text regions within an image and then employs Optical Character Recognition (OCR) to extract multilingual text from these areas. The extracted text characters are then transformed into visual shapes using a UTF-8 based text encoder and cross-attention. Recognizing that OCR may sometimes produce inaccurate results in real-world scenarios, we have developed two innovative methods to enhance the robustness of our model. By integrating text character priors with the low-resolution text images, our model effectively guides the super-resolution process, enhancing fine details within the text and improving overall legibility. The superior performance of our model on both the TextZoom and TextVQA datasets sets a new benchmark for STISR, underscoring the efficacy of our approach.

\section{Introduction}
\label{sec:intro}

\begin{figure}[t]
    \centering
    \includegraphics[width=1.0\linewidth]{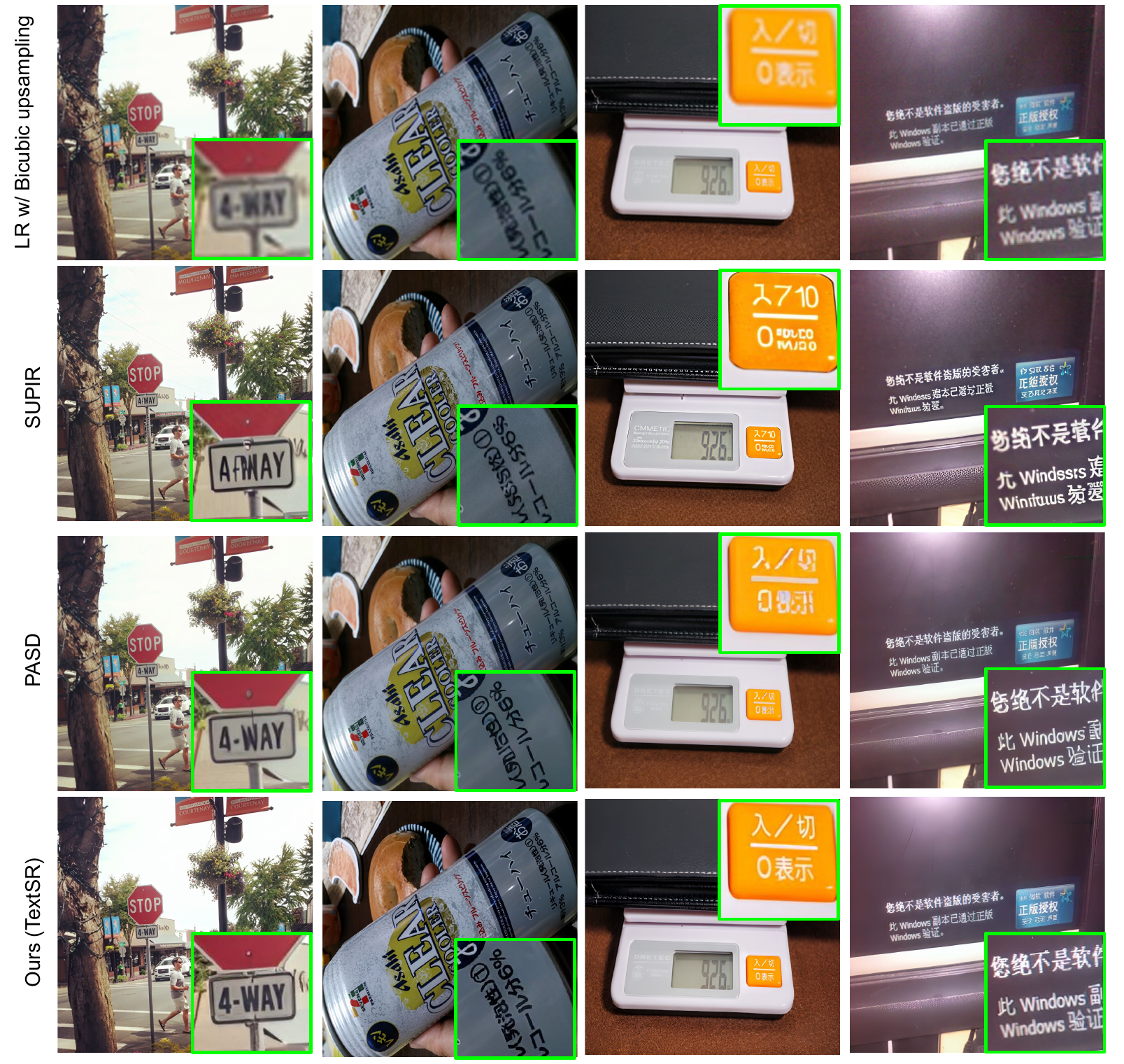}
    \caption{
    % Current state-of-the-art Super-Resolution model SUPIR \cite{Yu_2024_CVPR} leverages Multimodal Large Language Model LLaVA to enhance the identification of details within images. However, the caption from LLaVA fails to provide useful information regarding the character shape details. Our solution uses text detector for localizing the texts and it combines text recognition with restoration to refine individual characters.
    Our single model, equipped with multilingual character-to-shape diffusion priors, can super resolve low-resolution images in various languages and enhance both visual quality and legibility of text. This capability is lacking in current state-of-the-art super-resolution models (e.g., SUPIR~\cite{Yu_2024_CVPR}, PASD~\cite{yang2023pixel}).
    }
    \label{fig:teaser}
    \vspace{-6mm}
\end{figure}

% Image Super-Resolution (SR) involves recovering a high-resolution (HR) image from its low-resolution (LR) counterpart. SR can leverage external priors to ``complete'' the missing or degraded image details. For instance, given a pixelated low-resolution image of a pet, a model can generate realistic fur textures to enhance the image while maintaining fidelity to the original content. Another example is face SR~\cite{Li_2018_ECCV,liu2019reference,Li_2020_CVPR,Varanka_2024_CVPR,li2022learning}, where the model can use additional images of that same individual to faithfully restore facial features. SR solutions strike a balance between perceptual quality and content fidelity. This means that reliable SR models should generate believable image details that satisfy certain fidelity criteria while also appearing visually pleasing and natural.
% , even if those details were not in the original low-quality image, the results are considered acceptable and even desirable.
Image Super-Resolution (SR) aims to reconstruct a high-resolution (HR) image from a low-resolution (LR) input. By utilizing external priors, SR models can fill in missing or degraded image details. For example, given a pixelated low-resolution image of a pet, an SR model can generate realistic fur textures, enhancing the image while preserving the original content. Effective SR solutions achieve a balance between perceptual quality and content fidelity. This ensures that reliable SR models generate plausible image details that meet specific fidelity criteria while also appearing visually pleasing and natural.

Numerous SR approaches learn from large-scale datasets and encode the dataset priors in increasingly large neural networks to improve image details:
% address the encoding of external priors by learning through large-scale neural network models: 
from early regression-based Convolutional Neural Networks (CNNs)~\cite{dong2014learning,dong2015image,Kim_2016_CVPR,Zhang_2018_CVPR,Tai_2017_CVPR,delbracio2021projected}, to generative adversarial networks (GANs)~\cite{wang2018esrgan,Zhang_2021_ICCV,wang2021real,liang2021swinir}, and more recently, diffusion models (DMs)~\cite{saharia2022image,yang2023pixel,wang2024exploiting,delbracio2023inversion,Yu_2024_CVPR,Wu_2024_CVPR}.
However, larger model size does not necessarily translate to improved inductive biases in scene text SR. Such models can struggle to identify specific image areas that require text-focused restoration, leading to inconsistencies, hallucinated textures, and other inaccuracies in the generated images (see Fig.~\ref{fig:teaser}). These deviations from the original input can negatively impact the overall perceived visual quality and legibility of text.

Encoding the external priors with Multimodal Large Language Models (MLLMs) is a more sophisticated and efficient method compared to simply increasing the size of the models.
Building upon the foundational Latent Diffusion Models~\cite{Rombach_2022_CVPR}, approaches such as SUPIR~\cite{Yu_2024_CVPR}, PASD~\cite{yang2023pixel}, SeeSR~\cite{Wu_2024_CVPR}, SPIRE~\cite{qi2024spire} have introduced semantic prompts to steer the image enhancement process.
% Fig.~\ref{fig:teaser} left shows the processing steps of the SUPIR~\cite{Yu_2024_CVPR}
For example, in SUPIR~\cite{Yu_2024_CVPR}, the image caption is first extracted using the LLaVA~\cite{Liu_2024_CVPR}. Then, the LLaVA text and the low-resolution image are both fed to the SUPIR model to generate the high-resolution image.
% However, the SUPIR model fails in this specific example. First, it fails to accurately pinpoint the text regions that require restoration because the cross-attention mechanism for measuring text-image similarity does not effectively generalize to scene text images. Second, the method lacks the necessary components for mapping characters to their corresponding shapes.
Despite these advancements, MLLM-based approaches to STISR face challenges. Image-level text prompts generated by MLLMs lack the precision to guide the STISR model in identifying specific text regions for restoration. Furthermore, these prompts often fail to describe the content within text regions, particularly when dealing with foreign languages (see Fig.~\ref{fig:mllm_failed}).
% (e.g., the prompt ``The Arabic text on the banner is written in a bold and clear style'' provides nothing useful for reconstructing the text contents in the pixel space).

It is intuitive to adopt both OCR detection and recognition models to help with the aforementioned challenges faced by the MLLM-based methods. Methods such as \cite{textzoom,quan2020collaborative,zhao2021scene,wang2019textsr,xu2017learning,ma2023text,Ma_2022_CVPR,zhao2022c3stisrscenetextimage,Zhu_Zhao_Fang_Xue_2023} learn SR models that are applied solely to the text regions, assuming the off-the-shelf text detectors. Additionally,  \cite{Ma_2022_CVPR,zhao2022c3stisrscenetextimage,Zhu_Zhao_Fang_Xue_2023,liu2023textdiff,noguchi2024scene} utilize the text priors generated by the OCR recognition models. While these existing methods have significantly enhanced the legibility of restored text images, their reliance on datasets predominantly containing English text crops (e.g., TextZoom~\cite{textzoom}) limits their applicability. Although language-specific solutions exist (\cite{Zhang_2024_CVPR,Li_2023_CVPR} for Chinese characters), a single unified generalizable STISR model capable of handling multiple languages remains absent.

In this paper, we present the first STISR paradigm utilizing multilingual priors acquired from extensive text detection and transcription (TDT) datasets. In contrast to previous study, which primarily focused on single languages, our single model can effectively super resolve text images in over five languages (see Fig.~\ref{fig:teaser}).

% Scene text images present distinct challenges for super-resolution (SR) algorithms, as illustrated in the above example. These challenges differ from those encountered in general SR applications. We hypothesize that by incorporating inductive biases that promote simpler explanations, we can improve the generalization capabilities of these models when faced with unseen data. Specifically, we simplify the text localization using a text detection algorithm, and explicitly learn a SR model solely applied to the text regions. For this region-focused model, we consider both the image priors and the multilingual character-to-shape priors.

We name our model \textit{TextSR}, in short for Text Super-Resolution. 
% It learns LR image priors and multilingual character-to-shape priors in an alternating process. 
% Using a unique UTF-8 representation, 
The use of UTF-8 coding as the intermediate text representation is the core technique behind its ability to solve multilingual tasks.
% our model can be applied to various languages.
This representation allows for shared codebooks between languages with common characters, such as Chinese and Japanese. To train our model, we used a large and diverse dataset consisting of over 18 million text crops in more than five languages, gathered from seven text detection and transcription datasets.
% The size of our training dataset is significantly larger than those used in previous studies~\cite{textzoom,Ma_2023_ICCV}.
The diverse text images in multiple languages and the significantly larger size of our training dataset distinguishes our data from previous works~\cite{textzoom,Ma_2023_ICCV}.
To handle discrepancies between training and real-world input during model inference, we have introduced two innovative techniques, specifically designed to manage the challenges posed by noisy and inaccurate text recognition results. The results of our model on TextZoom~\cite{textzoom} and TextVQA~\cite{Singh_2019_CVPR} set a new state-of-the-art, validating the effectiveness of our approach.
Our contribution can be summarized as follows:

\begin{itemize}
    % \item We introduce a comprehensive framework (Fig.~\ref{fig:teaser} right) designed to tackle the difficulties of both text localization and scene text image super-resolution.

    \item We present a unified approach for modeling the multilingual character-to-shape diffusion priors. Our single STISR model can accommodate diverse languages.
    
    \item We explore the practical implications of text priors, focusing on how they function in the presence of photo capture and processing degradations, as well as inaccuracies stemming from external OCR sources.

    % \item We present a more comprehensive evaluation for the STISR under the SR framework, allowing comparison to the general SR methods. Vice versa, the proposed evaluation can be used to evaluate the unwanted visual artifacts introduced in the general SR models.
    
    \item We establish a new SOTA in both the region-based STISR evaluation (TextZoom) and holistic image evaluation (TextVQA).

\end{itemize}
\section{Related Work}
\label{sec:related}

\noindent\textbf{Image Super-Resolution} (SR), or single image super-resolution (SISR), is a computer vision task that enhances an image based on its low-resolution (LR) counterpart. Early SR methods mainly utilized convolutional neural networks~\cite{dong2016srcnn,lim2017edsr}. Then, the field was dominated by the transformer-based approaches, including SwinIR~\cite{liang2021swinir}, MAXIM~\cite{Tu_2022_CVPR}, and UFormer~\cite{Wang_2022_CVPR}, as well as adversarial-based methods like BSRGAN~\cite{Zhang_2021_ICCV} and Real-ESRGAN~\cite{wang2021realesrgantrainingrealworldblind}. More recently, the diffusion models (DMs)~\cite{NEURIPS2020_4c5bcfec} have gained prominence due to their ability to generate photorealistic images through iterative restoration \cite{whang2022deblurring,saharia2022image,li2022srdiff,saharia2022palette,delbracio2023inversion,ren2023multiscale}.

Building upon these advancements, recent state-of-the-art SR techniques leverage Multimodal Large Language Models (MLLMs) to further enhance the ability of diffusion models to recover intricate visual details. This represents a departure from methods that directly integrated image completion data into expanded model weights. By supplying textual cues to the diffusion model, the MLLM empowers it to generate plausible and visually consistent details, even those that may not be discernible in the low-resolution input.
For instance, the SUPIR model~\cite{Yu_2024_CVPR} utilizes LLaVA~\cite{Liu_2024_CVPR} to generate captions for low-resolution images. The PASD model~\cite{yang2023pixel} relies on YOLO~\cite{Redmon_2016_CVPR} and BLIP2~\cite{li2023blip} to provide object and semantic information. Furthermore, both SeeSR~\cite{Wu_2024_CVPR} and SPIRE~\cite{qi2024spire} employ a frozen semantic prompting branch in conjunction with a trainable degradation-aware prompt extractor.

\begin{figure}[t]
    \centering
    \includegraphics[width=1.0\linewidth]{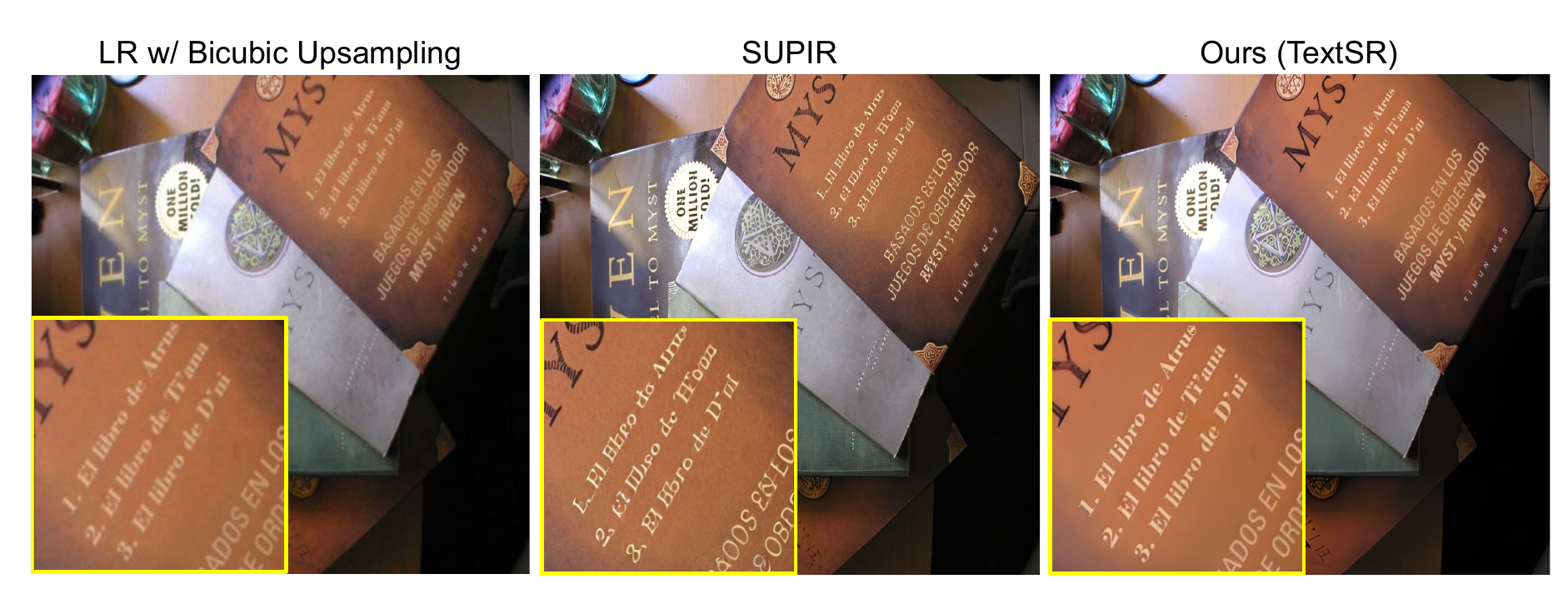}
    \caption{MLLMs were unable to accurately describe text content. Our method takes OCR detection results and recognized text contents. SUPIR~\cite{Yu_2024_CVPR} takes the full images and the following LLaVa~\cite{Liu_2024_CVPR} prompts: \textit{The image features a wooden table with a collection of books and pamphlets placed on top of it. There are three books in total, with one book being larger and positioned in the center of the table, while the other two books are smaller and located on the left and right sides of the table. 
    % The books are arranged in a way that they are easily accessible and visible to the viewer. The scene gives off a sense of organization and a focus on the books as the main subject.
    }}
    \label{fig:mllm_failed}
    \vspace{-4mm}
\end{figure}

MLLM-based SR methods struggle with scene text images (see Fig.~\ref{fig:mllm_failed}) because captioning may overlook non-salient text, cross-attention mechanisms require extensive training, and the model may confuse text with textures. Therefore, we propose an OCR-guided diffusion model to address these shortcomings. The framework resembles MLLM-based SR, but with key distinctions: the localization of texts relies on OCR detection rather than learned cross-attention; the image captioning utilizes text recognition instead of an MLLM; finally, our STISR's diffusion is localized solely to text regions.

\label{sec:method}
\begin{figure*}[t]
    % https://docs.google.com/drawings/d/1Diw_B1nTvI-AsBepSqvVNrtPbLASty6pEboQFVlz9RU/edit?usp=sharing
    \centering
    \includegraphics[width=0.95\linewidth]{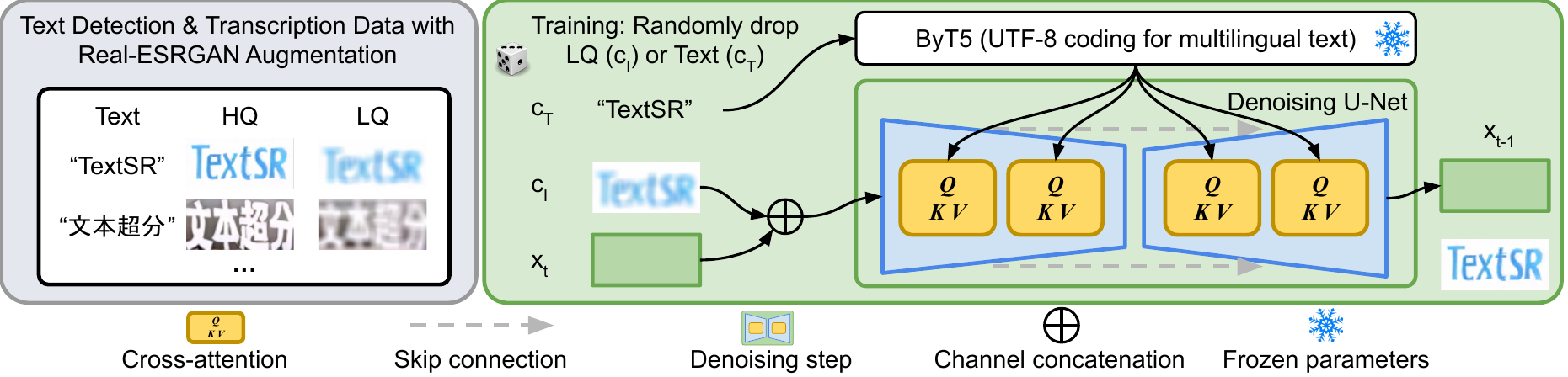}
    \vspace{-2mm}
    \caption{Model architecture and training pipeline.}
    \label{fig:architecture}
    \vspace{-4mm}
\end{figure*}

\vspace{1mm}
\noindent\textbf{Scene Text Image Super-Resolution} (STISR) focuses on enhancing the resolution of text-specific areas within images, such as individual words or sentences. The TextZoom dataset~\cite{textzoom}, with its diverse collection of low- and high-resolution image pairs, was instrumental in advancing this field. Subsequent studies, such as those by \cite{quan2020collaborative} and PCAN~\cite{zhao2021scene}, proposed methods for separating text images based on frequency bands, emphasizing the prediction of high-frequency information to enhance text clarity. The integration of text recognition models offered another avenue for STISR improvement. \cite{wang2019textsr} and \cite{xu2017learning} introduced the use of text recognizer loss as a perceptual loss to guide STISR model training. In comparison, other approaches, such as TPGSR~\cite{ma2023text} and TATT~\cite{Ma_2022_CVPR}, utilize OCR texts as additional input, employing transformer-based modules for semantic guidance during text reconstruction. Building on this, C3-STISR~\cite{zhao2022c3stisrscenetextimage} and DPMN~\cite{Zhu_Zhao_Fang_Xue_2023} further augmented text priors with additional modalities, such as text skeletons.

Recent advancements in STISR have seen the emergence of diffusion-based methods. TextDiff~\cite{liu2023textdiff} presented a diffusion-based model for predicting residuals. In a separate development, DCDM~\cite{noguchi2024scene} proposed the use of a diffusion model's text-to-image capabilities for the synthesis of low-quality and high-quality training data pairs.
Our approach is the first in STISR research to introduce a dual-conditioned generative model that models multilingual character-to-shape priors. We achieve this by leveraging text priors as an additional multimodal input during the diffusion process. This is done by adapting the cross-attention~\cite{NIPS2017_3f5ee243}, originally introduced in latent diffusion models~\cite{Rombach_2022_CVPR}.
In addition to introducing a new modeling approach, we suggest a practical STISR evaluation (Sec.~\ref{sec:results:text_vqa}) that aligns with general SR evaluation methods. This alignment enables a fair comparison with state-of-the-art DM-based methods \cite{wang2024exploiting,Yu_2024_CVPR,yang2023pixel}.

\vspace{1mm}
\noindent\textbf{Multilingual Scene Text Image Super-Resolution} is a natural extension of the STISR, requiring the handling of the restoration tasks of multilingual text images. \cite{Li_2023_CVPR} and \cite{Zhang_2024_CVPR} developed codebooks for text super-resolution tasks. \cite{Li_2023_CVPR}'s codebook encompasses 6,736 simplified Chinese characters, while \cite{Zhang_2024_CVPR} includes 7,934. Our approach employs a compact vocabulary of 259 characters. We leverage the ByT5 encoder~\cite{xue-etal-2022-byt5} for subsequent feature extraction. This approach enables a single model to efficiently support multiple languages by facilitating the sharing of codebooks across different languages.
Fig.~\ref{fig:char_to_shape_priors} illustrates our learned multilingual character-to-shape priors.

% \noindent\textbf{Virtual Text Rendering} is a task to render text images from text contents and pre-specified layout structure. Methods such as ~\cite{zhu2024visual}, \cite{NEURIPS2023_8951bbdc}, \cite{liu-etal-2023-character}... (Note \cite{liu-etal-2023-character} paper mentioned ByT5, so using ByT5 may not be novel, read this paper in details for writting Sec.\ref{sec:related:text_modeling}). TextDiffuser\cite{NEURIPS2023_1df4afb0,chen2023textdiffuser} represented the layout using character priors.

% \noindent\textbf{Style-transfer of Handwritten Text} is a task to synthesize new style of text image based on an source image, and a handwritten style label. Internally, it assumes the text contents of the source image is also available. There are many GAN-based methods to approach the problem. For example, \cite{Fogel_2020_CVPR} presented a ..., \cite{Kang_2020_WACV} introduced a ..., \cite{luo2022slogan} proposed to .... Due to the observation in \cite{NEURIPS2021_49ad23d1} that GAN models were leading less diversified results, hard-to-train, and easy-to-collapse, \cite{Zhu_2023_CVPR} present the first study in handwritten text stylizing using diffusion models \cite{NEURIPS2020_4c5bcfec}.

% \input{tabs/t2i_tasks}

\section{Methodology}

The task of holistic text image super-resolution involves creating a function, denoted as $f$, that enhances the legibility of an input image containing text. Given an input image, ${I}$, the function produces an output image, ${I}'=f({I})$, that is sharper, cleaner, and overall easier to read.
Recent diffusion-based methods for restoration such as the one proposed by \cite{wang2024exploiting}, could be applied to implement the function $f$. Despite claims of strong performance in text restoration, we found that these methods often produce textures that lack fidelity to the underlying characters (see Fig.~\ref{fig:text_vqa}).

Our proposed solution circumvents the direct approach to function $f$, opting instead to decompose it into a series of region-based restoration problems.
Formally, we use a text detector~\cite{gupta2016synthetic} capable of identifying the spatial locations of all textual elements within a given image. We can represent the detector's output as $D({I})=\{\theta_1,\dots,\theta_N\}$, where each $\theta_i \in \mathbb{R}^{2\times 3} (1\leq i\leq N)$ corresponds to a $2\times 3$ affine transformation matrix. This matrix encapsulates the cropping operation necessary to isolate the $i$-th text region within the image: ${T}_i = \Phi({I}, \theta_i)$, where $\Phi$ denotes the general affine transformation.
Our objective is to learn a restoration function $g$, a Scene Text Image Super-resolution (STISR) model, that enhances the legibility of a text region. This function transforms a cropped text line ${T}_i$ into a clearer version, represented as ${T}_i'=g({T}_i)$.
The final SR result can be obtained by iterative application of the function $g$: $I'=f(I) \oplus \Phi({T}_1', \theta_1^{-1}) \oplus \dots \oplus \Phi({T}_N', \theta_N^{-1})$, where $\theta_i^{-1}$ denotes the inverse affine transformation matrix, $\oplus$ is the blending operation, and $I'$ is the final output.

The effectiveness of our proposed solution hinges on how we model $g$. In the following, we present our diffusion-based solution within the context of $g$, beginning with the fundamentals of diffusion-based restoration methods in Sec.~\ref{sec:method:preliminaries}. Then, Sec.~\ref{sec:method:training} outlines our approach to multilingual text prior modeling. Finally, in Sec.~\ref{sec:method:inference}, we detail the adaptation of our solution to practical scenarios where precise text prompts were not available.

\subsection{Preliminaries of Diffusion Models}
\label{sec:method:preliminaries}

Denoising Diffusion Probabilistic Models (DDPMs), as introduced in~\cite{ho2020denoising}, are a type of generative models that learn to generate samples from the underlying data distribution. This is achieved by progressively denoising images that have been corrupted with Gaussian noise. 
Starting with a pure realization of Gaussian noise, the iterative denoising process is typically performed using a U-Net architecture~\cite{ronneberger2015u}, trained to predict the Gaussian noise $\epsilon$ present in a noisy image $x_t$ at a given timestep $t$. This prediction,  denoted as $\epsilon_{\theta}(x_t, t)$, is trained by minimizing the loss function
\begin{equation}
    L(\theta) = \mathbb{E}_{t, x_0, \epsilon} ||\epsilon - \epsilon_{\theta}(x_t, t)||_2^2,
\end{equation}
where $t$ is sampled uniformly from $\{1, \dots, T\}$, and $\epsilon \sim \mathcal{N}(0, Id)$.
To speed up inference, it is common to employ the DDIM sampling technique~\cite{song2020denoising}, which allows for skipping denoising steps during this stage. This inference optimization has been notably used in \cite{dhariwal2021diffusion,nichol2021improved}.

In diffusion-based image restoration models, the U-Net can receive information about the low-quality image $c_I$ by concatenating the condition with the input $x_t$ along the channel dimension (see Fig.~\ref{fig:architecture}). This combined input is then processed by the U-Net, denoted as $\epsilon_{\theta}(x_t, t, c_I)$.

\subsection{Training with Multilingual Texts}
\label{sec:method:training}

Leveraging the advancements in text-to-image generation~\cite{Rombach_2022_CVPR}, we introduce a model architecture tailored for the specific multilingual STISR task. The proposed model's key capability lies in its ability to leverage both the low-quality image $c_I$ and the text recognition results $c_T$ which are represented as a sequence of UTF-8 characters.

The model architecture, as shown in Fig.~\ref{fig:architecture}, is trained on triplets of low-resolution (LR) images, high-resolution (HR) images, and corresponding ground-truth text. The model can learn to condition on either or both modalities through random dropout of the text and image.
Our sampling process, unlike those in typical diffusion models, utilizes residual images derived by subtracting the LR image from the HR. This method helps to prevent the model from generating significant hallucinations.

\vspace{1mm}
\noindent\textbf{Multilingual Character Representations.}
The ground-truth text undergoes a series of transformations: conversion to UTF-8 byte format, assignment of unique token IDs (encompassing 256 UTF-8 characters along with special tokens PAD, EOS, and UNK), and subsequent mapping to associated UTF-8 character embeddings. Following this, a Transformer neural network architecture~\cite{NIPS2017_3f5ee243} is employed to extract lower-level text features.
% To support multilingual in STISR, we focus on N-gram language models with $N<4$. Unigrams suffice for Latin characters (e.g., 'a'-'z'), while trigrams are necessary for languages like Chinese, which use three UTF-8 characters per symbol.
% ByT5~\cite{xue-etal-2022-byt5} aligns with our needs as it offers semantic understanding in addition to character priors.
ByT5~\cite{xue-etal-2022-byt5} fulfills our requirements because it uses UTF-8 text representation and sequence-to-sequence translation as the primary training task to learn character priors. Additionally, ByT5 supports over 100+ languages, encompassing the languages relevant to our project.
We employ the initial two text encoder layers of ByT5, as the deeper encoder layers and decoder are more aligned with machine translation tasks and thus omitted. Throughout the training of the diffusion model, the ByT5 encoder layers remained frozen. The ByT5 layers produce an output denoted as $\tau(c_T)\in\mathbb{R}^{M\times d_{\tau}}$, where $M$ represents the UTF-8 string sequence length and $d_{\tau}$ represents the intermediate dimension.

\vspace{1mm}
\noindent\textbf{Conditioning Mechanism.}
We follow \cite{Rombach_2022_CVPR} which introduced the cross-attention mechanism~\cite{NIPS2017_3f5ee243} to diffusion models: $\text{Attention}(Q, K, V)=\text{softmax}(\frac{QK^T}{\sqrt{d}})\cdot V$, where $Q=W_Q^{(i)} \cdot \varphi(x_t), ~ K=W_K^{(i)}\cdot \tau(c_T), ~ V=W_V^{(i)}\cdot \tau(c_T)$, $\varphi(x_t)$ denotes the flattened U-Net intermediate feature map, $W_Q^{(i)}$, $W_K^{(i)}$, $W_V^{(i)}$ are the parameters for the cross-attention layers.
To explain the above equations: the cross-attention mechanism allows the U-Net to analyze how similar image features are to character features. This similarity assessment allows the model to apply what it knows about character shapes to the image features. For instance, in STISR, the model can differentiate between easily confused characters like lowercase I (`i'), lowercase L (`l'), and number one (`1') by leveraging this character knowledge produced by an OCR model.
We denote our final U-Net with LR condition $c_I$ and text character condition $c_T$ using $\epsilon_{\theta}(x_t, t, c_I, c_T)$.

\subsection{Inference with OCR Texts}
\label{sec:method:inference}

Conditioning the model on ground-truth text characters (Sec.~\ref{sec:method:training}) during training offers several advantages. Primarily, it guarantees the model receives accurate text information, potentially enabling it to learn basic character-level priors. Additionally, it eliminates the dependence on a real OCR recognizer, simplifying the process for training (no need to train multiple STISR models to adapt to different OCR models). However, a notable drawback is that the trained model remains unaware of the performance of the OCR model at inference time.
In what follows, we introduce different approaches to adapt text-based models, originally trained on precise ground-truth characters, to handle the uncertainties introduced by imprecise OCR outputs during inference.

\subsubsection{Classifier-Free Guidance with Dual Conditions}
\label{sec:method:inference:cfg}

Fig.~\ref{fig:architecture} illustrates our training process where ground-truth text annotations are randomly dropped.
% This approach can be interpreted as training two models with a shared backbone in an alternating fashion:
This specifically designed random dropout allows for the co-training of two models that are closely related to the STISR.
\begin{itemize}
    \item $\epsilon_{\theta}(x_t, t, c_I, \varnothing)$: an image-only-conditioned model consistently receives empty text $\varnothing$ as input.
    \item $\epsilon_{\theta}(x_t, t, c_I, c_T)$: an image-text-conditioned model that takes both the LR image and ground-truth text.
\end{itemize}

During inference, we utilize classifier-free guidance~\cite{ho2022classifierfreediffusionguidance,Brooks_2023_CVPR,qi2024spire} to balance image-guided diffusion and text-guided diffusion controlled by OCR recognized texts. Our primary focus is on addressing the challenges posed by using potentially unreliable OCR input. The $\omega$ in Eq. (2) serves as a trade-off term for fidelity and quality, while this trade-off is rarely studied in the previous STISR models.
A high $\omega$ value allows the model to better capture plausible details in character shapes when the text signal is clean and precise. However, when the text is generated by an imprecise OCR model, an $\omega$ value less than 1 enables the model to rely more on pixel information to reconstruct the output.
% Drawing inspiration from the techniques presented in \cite{Brooks_2023_CVPR, qi2024spire}, which leverage classifier-free guidance~\cite{ho2022classifierfreediffusionguidance} for instruction-guided image editing, we introduce a new formulation. Our approach integrates classifier-free diffusion guidance into two distinct terms, effectively balancing their contributions to achieve enhanced STISR outcomes.
% The parameter $\omega$ in Eq.~\ref{eq:cfg} is adjusted after the denoising model has been trained using ground-truth texts (Sec.~\ref{sec:method:training}). This parameter balances the OCR control and image fidelity.

\begin{equation}
\label{eq:cfg}
\begin{split}
    &\tilde \epsilon_{\theta}(x_t,t,c_I,c_T) \\
    =
    &\underbrace{(1-\omega)\cdot\epsilon_{\theta}(x_t,t,c_I,\varnothing)}_{\text{LR-Guided}}
    +
    \underbrace{\omega\!\cdot\!\epsilon_{\theta}(x_t,t,c_I,c_T)}_{\text{OCR-Guided}}
\end{split}
\end{equation}

\subsubsection{Iterative OCR Conditioning}
\label{sec:method:inference:iterative}

The application of noisy text recognition results, due to the impact of image quality on OCR performance, can be problematic. However, our STISR model offers a potential solution to this issue.
Let us represent an OCR model as $\psi(c_I)$ and our STISR model as $g(c_I, c_T)$ (i.e., the DDIM inference using the denoiser Eq.~\ref{eq:cfg}). We propose an iterative approach that alternates between two steps: (1) enhancing OCR accuracy using the restored image, and (2) leveraging improved OCR results to further refine image restoration. Formally, given $R$ the number of iterations, and $g^{(R)}$ our STISR restoration function using $R$- iterations:

\begin{itemize}
    \item $R=0$: $g^{(0)}(c_I)=g(c_I, \psi(c_I))$; i.e., we use the OCR text from the LR image to guide the restoration model.
    
    \item $R=1$: $g^{(1)}(c_I)=g(c_I, \psi(g(c_I, \varnothing)))$; in addition to the above approach, we use the image-only model $g(c_I, \varnothing)$ to preprocess the OCR input image.
    
    \item $R>1$: $g^{(R)}(c_I)=g(c_I, \psi(g^{(R-1)}(c_I)))$.
\end{itemize}

\vspace{1mm}
Setting $R\!=\!1$ significantly improves the STISR performance (see Sec.~\ref{sec:results:text_zoom}) as compared to $R\!=\!0$. This suggests that when dealing with degraded images, the input to the OCR system can be enhanced by utilizing our image-only model $\epsilon_{\theta}(x_t, t, c_I, \varnothing)$.
As compared to the previous work \cite{ma2023text} which also proposed ``iterative refinement'', the underlying difference is that our method additionally updates the text prior $c_T$ and ours is a diffusion-based method.
\section{Experiments}
\label{sec:results}

% We begin by introducing the training/evaluation datasets. Then, we detail the implementation of our model. Next, we analyze our model's STISR performance using the TextZoom laboratory environment dataset~\cite{textzoom}. Finally, we present a comparative assessment using the widely adopted TextVQA~\cite{Singh_2019_CVPR} benchmark and contrast our results with more comprehensive image SR baselines.
% First, we introduce the data used for training and evaluation. Next, we detail our implementation and showcase the learned multilingual character-to-shape priors. We then present a comparative assessment using the TextVQA benchmark~\cite{Singh_2019_CVPR} and contrast our results with more comprehensive image SR baselines. Finally, we  analyze our model's performance using the TextZoom latin dataset~\cite{textzoom}.

\noindent\textbf{Data for Training and Evaluation.}
Our STISR model was trained on a mixed training collection comprising seven text detection and transcription (TDT) datasets: HierText~\cite{Long_2022_CVPR} (8,281 samples), TextOCR~\cite{singh2021textocr} (21,778 samples), VinText~\cite{Nguyen_2021_CVPR} (1,200 samples), DenseText~\cite{densetext} (1,038 samples), ICDAR19 LSVT~\cite{sun2019icdar} (30,000 samples), ICDAR19 MLT~\cite{nayef2019icdar2019robustreadingchallenge} (10,000 samples), and TotalText~\cite{ch2017total} (1,255 samples). 
The above datasets only include clean text images with detection and transcription annotations. Hence, we introduced low-quality images into these datasets using the Real-ESRGAN pipeline~\cite{wang2021realesrgantrainingrealworldblind}. Each image was duplicated 20 times then each had random high-order Real-ESRGAN degradation operations applied. Next, we used the ground-truth detection annotations to extract text regions. We then filtered out any regions with a height outside the range of 16 to 512 pixels. This process resulted in the following number of crops: 3.8M from HierText, 7.8M from TextOCR, 435K from VinText, 364K from DenseText, 4.4M from LSVT, 1.7M from MLT, and 195K from TotalText.

To evaluate our model, we used the TextVQA~\cite{Singh_2019_CVPR} and TextZoom~\cite{textzoom} datasets. TextZoom simulates a controlled environment where low- and high-quality images are captured at varying camera focal lengths and subsequently aligned using registration methods. In comparison, TextVQA provides a comprehensive assessment with its samples, including both OCR~\cite{singh2021textocr} and VQA (Visual Question Answering)~\cite{Singh_2019_CVPR} annotations.

\vspace{1mm}
\noindent\textbf{Implementation Details.}
% Config: google3/experimental/research/vision/tenex/textsr/configs/public_mixed_l5_c32_g4_48x480.py
In our text modeling, we employ line abstraction to enhance efficiency and leverage OCR context. During the training phase, input text images are standardized to a height of 48 pixels. We then either slice or pad these images to ensure a uniform dimension of $48\!\times\!480$ pixels. Subsequently, the textual content within these images is converted into UTF-8 codes. For feature extraction, we employ the  \textsc{ByT5-Base} \cite{xue-etal-2022-byt5}, distinguished by its 1,536 embedding dimensions, 3,968 intermediate dimensions, and 12 attention heads. However, for our specific purposes, we utilize only the initial two layers of the text encoder.
Our U-Net denoiser begins with 32 channels and uses five residual down/up block pairs. Each pair has 2 down layers and 3 up layers, with channel multipliers of 1, 2, 4, 8, 8. When processing a $48\!\times\!480$ text image, the down blocks downsample the image to various feature map sizes. Cross-attention~\cite{NIPS2017_3f5ee243} introduces the ByT5 representations only to the $3\!\times\!30\!\times\!256$ and $1\!\times\!10\!\times\!256$ feature maps.

For inference, we utilize a 5-step DDIM sampling method~\cite{song2020denoising}. To ensure smooth transitions between the 16-pixel tile overlaps (line image's aspect ratio may exceeds 1:10), we employ a interpolation technique.
Our model training was executed using JAX~\cite{jax2018github} and TPUv5, with a batch size of 1,024 and a learning rate of 3e-4.
For TextVQA, we utilized an internal OCR with multilingual support for both detection and transcription. The OCR models we used have practical mistakes, reflecting real-world OCR performance. To generate results for complete images, we trained a Real-ESRGAN model ($f$)~\cite{wang2021realesrgantrainingrealworldblind} on the LSDIR dataset~\cite{li2023lsdir} and integrated it with our TextSR ($g$). More details are provided in the supplementary materials.
We performed additional fine-tuning for the TextZoom.

\begin{figure}[t]
    \centering
    \includegraphics[width=1.0\linewidth]{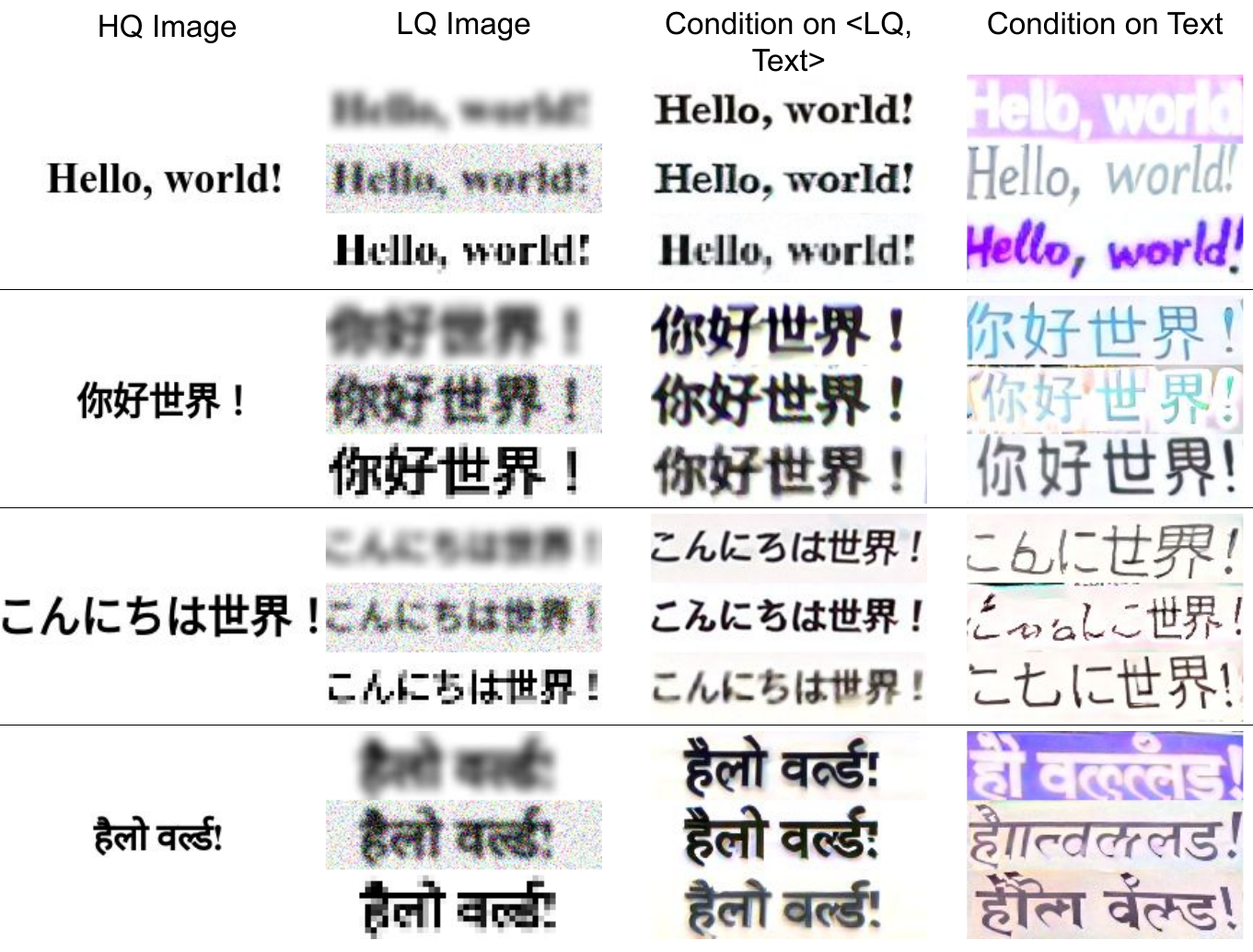}
    \caption{Qualitative results showing the learned multilingual character-to-shape priors (from top to down: English, Chinese, Japanese, and Hindi). The visualization was generated by setting $\omega$ to 10.0 for the $<$LQ, text$>$-conditioned model and 15.0 for the text-only model. For the text-only model, $c_I$ was set to $\varnothing$, and three random seeds were applied. The HQ images were degraded by blurring, adding noise, and downsampling to create the three LQ images for each language group.}
    \label{fig:char_to_shape_priors}
    \vspace{-4mm}
\end{figure}

\subsection{Multilingual Character-to-shape Priors}
Our model, trained on 18 million text crops, can predict character shapes across multiple languages, including English, Chinese, Japanese, and Hindi. This is demonstrated in Fig.~\ref{fig:char_to_shape_priors}, where high- and low-quality images are presented alongside our model's predictions using both low-quality images and text input, as well as text-only input. The results demonstrate that our model can produce character shapes solely based on text prompts, even without visual input. This capability, which we refer to as ``multilingual character-to-shape diffusion priors,'' could be particularly useful for text restoration tasks --- we quantitatively investigate the model in the following experiments.

\subsection{Results on the TextVQA}
\label{sec:results:text_vqa}

% \begin{figure*}[t]
%     \centering
%     \includegraphics[width=1.0\linewidth]{figs/TextVQAQualitativeResults.pdf}
%     \caption{Qualitative results on the TextVQA.}
%     \label{fig:text_vqa}
% \vspace{-5mm}
% \end{figure*}
\begin{figure*}[t]
    \centering
    \includegraphics[width=1.0\linewidth]{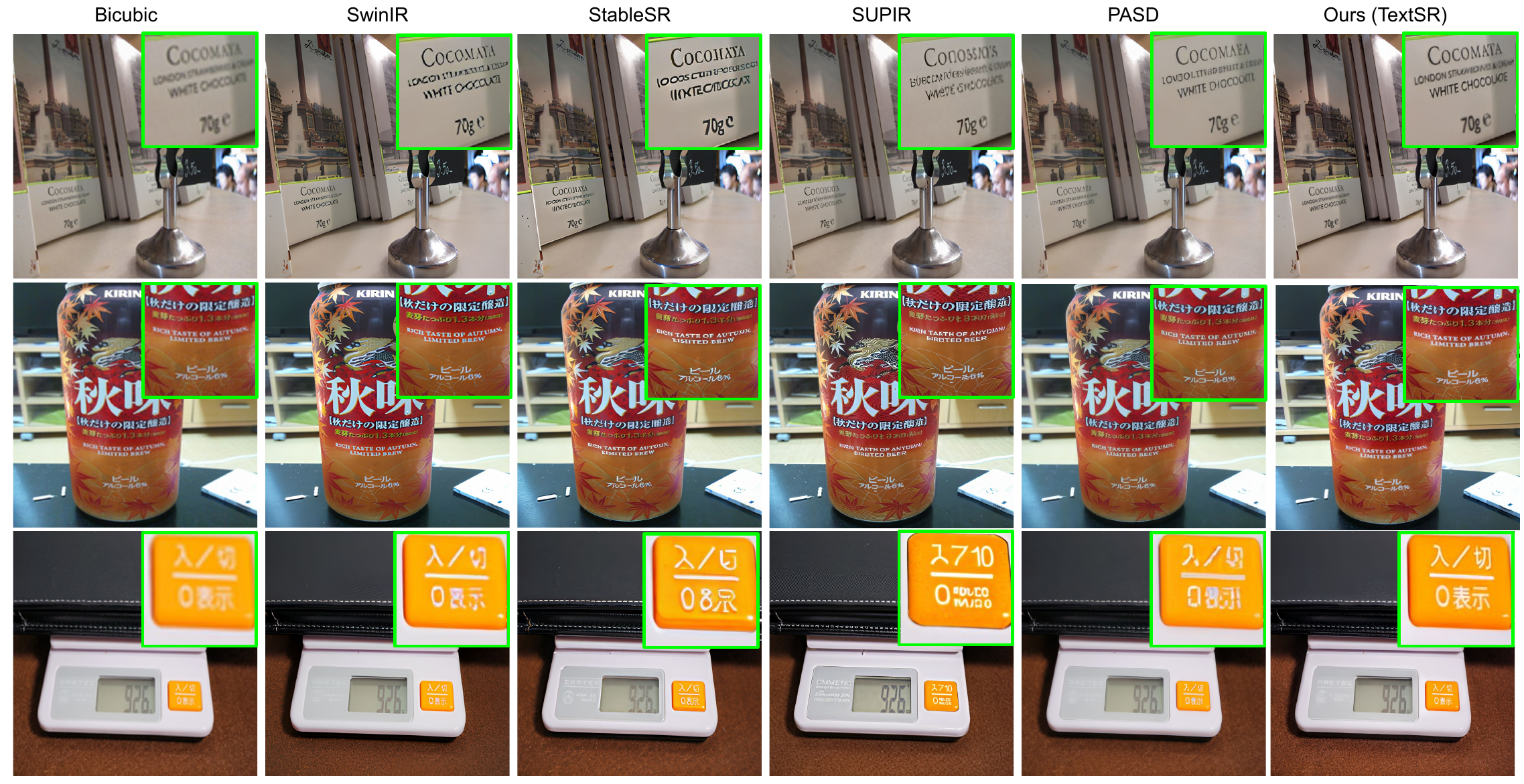}
    \caption{Qualitative results on the TextVQA. State-of-the-art methods were ineffective in the presence of foreign texts.}
    \label{fig:text_vqa}
    \vspace{-6mm}
\end{figure*}

% The TextZoom benchmark, while commonly used to evaluate STISR algorithms, has limitations that impact its applicability to real-world scenarios. Its primary focus on word-level texts with a 1:4 aspect ratio ($32\!\times\!128$) restricts the utilization of broader contextual information. Furthermore, the benchmark's reliance on word-level analysis poses challenges for languages like Chinese and Japanese, which lack clear word boundaries defined by spaces or symbols.
% In contrast, our method, prior to any adaptation or fine-tuning on TextZoom, utilizes a 1:10 ($48\!\times\!480$) inference engine on text line tiles. It preserves more OCR characters, maintains the original aspect ratio, and supports multilingual characters -- a feature unable to be evaluated by the TextZoom benchmark.
We introduce a new benchmark task leveraging the extensive TextVQA dataset~\cite{Singh_2019_CVPR}. Notably, evaluation is not confined to text regions; models must generate complete, full-resolution images. This facilitates direct comparison with state-of-the-art general SR methods~\cite{wang2024exploiting,Yu_2024_CVPR,yang2023pixel}. In this study, we used the 641 images from the initial 1,000 questions in the TextVQA validation set. These images were downscaled to $512\!\times\!512$ and $256\!\times\!256$ pixel resolutions. We then applied various super-resolution (SR) methods to upscale these images by factors of 2x and 4x, respectively, to achieve a final resolution of 1MP. 
To assess the quality of these upscaled images, we have incorporated the OCR accuracy as measured by the CRNN~\cite{shi2016end} provided in \cite{textzoom} to account for the subtle errors made by the various models. We have categorized OCR evaluation of the 641 images into three data buckets based on text size: 15,783 small text crops (height$<$32 px), 4,273 medium crops ($<$64 px), and 1,423 large crops ($\ge$64 px).
Additionally, we employed the GIT~\cite{wang2022git} model (Hugging Face ID ``microsoft/git-large-textvqa'') trained on the TextVQA dataset. This AI agent answered questions using the enhanced images. Our underlying assumption is that superior SR methods would produce higher-quality images, leading the AI agent to provide more accurate answers.
The VQA accuracy is adopted to measure the performance of different SR methods:
$\text{VQA Acc(\textit{ans})}=min\{\frac{\text{\#humans that said \textit{ans}}}{3}, 1\}$.

\begin{table}[t]
    \footnotesize
    \centering
    \setlength\tabcolsep{4pt}
    \begin{tabular}{clcccccccc}
        \toprule
            & \multirow{2}{*}{\textbf{Method}} & \multicolumn{3}{c}{\textbf{OCR Acc.}} & \multirow{2}{*}{\parbox{1cm}{\centering \textbf{VQA Acc.}}}\\
            & & \textbf{Small} & \textbf{Medium} & \textbf{Large} \\
        \midrule
            \multirow{9}{*}{4$\times$}
            & Bicubic
                & 1.7\% & 21.1\% & \second{40.6\%} & \second{33.2\%} \\
            & BSRGAN~\cite{Zhang_2021_ICCV}
                & 2.3\% & 20.0\% & 37.8\% & 31.0\% \\
            & Real-ESRGAN~\cite{wang2021real}
                & 2.4\% & 20.5\% & 37.7\% & 30.4\% \\
            & SwinIR~\cite{liang2021swinir}
                & 2.9\% & 22.4\% & 37.7\% & 31.5\% \\
            & StableSR~\cite{wang2024exploiting}
                & 3.2\% & 20.2\% & 35.6\% & 28.1\% \\
            & SUPIR~\cite{Yu_2024_CVPR}
                & 2.5\% & 18.2\% & 34.6\% & 28.9\% \\
            & PASD~\cite{yang2023pixel}
                & 4.7\% & 28.9\% & 39.2\% & 31.3\% \\
        \cline{2-6}
            & Ours (Real-ESRGAN)
                & \second{5.1\%} & \second{30.2\%} & 40.3\% & \first{33.9\%} \\
            & Ours (+TextSR)
                & \first{7.3\%} & \first{36.7\%} & \first{40.8\%} & 32.3\% \\
        \midrule
            \multirow{9}{*}{2$\times$}
            & Bicubic
                & 18.8\% & 48.1\% & \first{45.1\%} & \second{35.4\%} \\
            & BSRGAN~\cite{Zhang_2021_ICCV}
                & 16.9\% & 47.3\% & 41.7\% & 35.0\% \\
            & Real-ESRGAN~\cite{wang2021real}
                & 12.9\% & 44.5\% & 42.3\% & 34.6\% \\
            & SwinIR~\cite{liang2021swinir}
                & \second{20.0\%} & \second{48.5\%} & 41.9\% & \second{35.4\%} \\
            & StableSR~\cite{wang2024exploiting}
                & 4.4\% & 24.4\% & 34.9\% & 30.7\% \\
            & SUPIR~\cite{Yu_2024_CVPR}
                & 5.1\% & 26.4\% & 38.5\% & 30.7\% \\
            & PASD~\cite{yang2023pixel}
                & 4.7\% & 28.1\% & 39.1\% & 31.6\% \\
        \cline{2-6}
            & Ours (Real-ESRGAN)
                & 19.7\% & \second{48.5\%} & \second{42.5\%} & \first{35.6\%} \\
            & Ours (+TextSR)
                & \first{31.4\%} & \first{51.8\%} & 43.4\% & \second{35.4\%} \\
        \bottomrule
    \end{tabular}
    \caption{Results on the TextVQA Dataset. VQA answers were provided by the AI agent GIT~\cite{wang2022git}, while the OCR results were extracted by the CRNN~\cite{shi2016end} provided in \cite{textzoom}. The best numbers are highlighted in \first{blue}, followed by the second-best in \second{black}.}
    \label{tab:text_vqa}
\vspace{-5mm}
\end{table}

\vspace{1mm}
\noindent\textbf{Comparing to SOTAs.}
For comparison, we include the SOTA SR models implemented using general adversarial networks (BSRGAN~\cite{Zhang_2021_ICCV}, Real-ESRGAN~\cite{wang2021real}, SwinIR~\cite{liang2021swinir}), and diffusion models (StableSR~\cite{wang2024exploiting}, SUPIR~\cite{Yu_2024_CVPR}, PASD~\cite{yang2023pixel}). Bicubic upsampling is used as a strong baseline --- beating it in terms of text legibility in the restored image is not simple.
For our methods, we present Ours (Real-ESRGAN) which serves as the model $f$, as discussed in Sec.~\ref{sec:method}. Additionally, we present the final super-resolution results, referred to as Ours (TextSR) ($I'=f\oplus \dots$), which are the outcomes achieved by blending with Ours (Real-ESRGAN).

The results are shown in Tab.~\ref{tab:text_vqa}. 
We use the 4x results to describe our observations. Primarily, it is important to recognize that the majority of methods negatively impact the legibility of large text ($\ge$64 px). Ideally, our method should not affect large text. However, due to blending the results with the Real-ESRGAN model, the results were slightly affected. For DM-based StableSR~\cite{wang2024exploiting} and SUPIR~\cite{Yu_2024_CVPR}, there is a significant reduction in legibility with respect to OCR accuracy (Bicubic 40.6\%, StableSR 35.6\%, SUPIR 34.6\%), consequently leading to inaccurate VQA performance (Bicubic 33.2\%, StableSR 28.1\%, SUPIR 28.9\%). The supplementary materials and the qualitative examples in Fig.~\ref{fig:text_vqa} illustrate the reasons for this outcome.
In these examples, StableSR and SUPIR generate incorrect characters or even non-text textures.
These results indicate that these SOTA SR methods need better design to promote fidelity of text.
Our method stands out as the only one to achieve significant text quality improvements. Compared to the second-best method, PASD, we observed gains of +2.6 points (7.3\% vs. 4.7\%) on the small split and a substantial +7.5 point improvement (36.7\% vs. 28.9\%) on the medium split. For a clearer demonstration of these visual enhancements, which may not be fully captured by OCR/VQA accuracy metrics, we refer readers to the qualitative examples in our supplementary materials.

\begin{table*}[t]
    \footnotesize
    \centering
    \setlength\tabcolsep{5pt}
    \begin{tabular}{lcccccccccccc}
        \toprule
            \multirow{2}{*}{\textbf{Method}} & \multicolumn{4}{c}{\textbf{CRNN}~\cite{shi2016end}} & \multicolumn{4}{c}{\textbf{MORAN}~\cite{luo2019moran}} & \multicolumn{4}{c}{\textbf{ASTER}~\cite{shi2018aster}}\\
            & \textbf{Easy} & \textbf{Medium} & \textbf{Hard} & \textbf{Avg.} & \textbf{Easy} & \textbf{Medium} & \textbf{Hard} & \textbf{Avg.} & \textbf{Easy} & \textbf{Medium} & \textbf{Hard} & \textbf{Avg.} \\
        \midrule
            LR (Bicubic)                            & 36.4\% & 21.1\% & 21.1\% & 26.8\% & 60.6\% & 37.9\% & 30.8\% & 44.1\% & 64.7\% & 42.4\% & 31.2\% & 47.2\% \\
            HR*                                     & 76.4\% & 75.1\% & 64.6\% & 72.4\% & 91.2\% & 85.3\% & 74.2\% & 84.1\% & 94.2\% & 87.7\% & 76.2\% & 86.6\% \\
        \midrule
            TSRN~\cite{textzoom}                    & 52.5\% & 38.2\% & 31.4\% & 41.4\% & 70.1\% & 53.3\% & 37.9\% & 54.8\% & 75.1\% & 56.3\% & 40.1\% & 58.3\% \\
            TATT~\cite{Ma_2022_CVPR}                & 62.6\% & 53.4\% & 39.8\% & 52.6\% & 72.5\% & 60.2\% & 43.1\% & 59.5\% & 78.9\% & 63.4\% & 45.4\% & 63.6\% \\
            TATT+DPMN~\cite{Zhu_Zhao_Fang_Xue_2023} & 64.4\% & 54.2\% & 39.2\% & 53.4\% & 73.3\% & 61.5\% & 43.9\% & 60.4\% & 79.3\% & 64.1\% & 45.2\% & 63.9\% \\
            TextDiff~\cite{liu2023textdiff}         & 64.8\% & 55.4\% & 39.9\% & 54.2\% & 77.7\% & 62.5\% & 44.6\% & 62.7\% & 80.8\% & \first{66.5\%} & \first{48.7\%} & \first{66.4\%} \\
            TCDM~\cite{noguchi2024scene}            & 67.3\% & 57.3\% & \second{42.7\%} & 55.7\% & 77.6\% & 62.9\% & 45.9\% & 62.2\% & \second{81.3\%} & \second{65.1\%} & 50.1\% & 65.5\% \\
            DCDM~\cite{singhdcdm}                   & 65.7\% & 57.3\% & 41.4\% & 55.5\% & \second{78.4\%} & \first{63.5\%} & 45.3\% & \second{63.4\%} & \first{81.8\%} & \second{65.1\%} & 47.4\% & \second{65.8\%} \\
        \midrule
            Ours (\textsc{ImgOnly})                 & \second{67.8\%} & \second{57.8\%} & 41.8\% & \second{56.6\%} & 77.7\% & 61.9\% & \second{46.2\%} & 63.0\% & 80.8\% & 64.1\% & 48.3\% & 65.4\% \\
            Ours (\textsc{+Gt})*                     & 80.4\% & 77.6\% & 71.1\% & 76.6\% & 90.7\% & 84.9\% & 77.1\% & 84.7\% & 94.4\% & 87.2\% & 78.6\% & 87.2\% \\
            Ours (\textsc{+Iter. Aster})              & \first{69.9\%} & \first{60.0\%} & \first{43.9\%} & \first{58.7\%} & \first{78.7\%} & \second{63.2\%} & \first{47.1\%} & \first{64.0\%} & 81.0\% & 64.6\% & \first{48.7\%} & \second{65.8\%} \\
        \bottomrule
    \end{tabular}
    \caption{Results on the TextZoom Dataset. The table displays OCR accuracy rates achieved using various OCR recognizers. An asterisk (*) indicates methods that incorporate ground-truth information (HR utilizes ground-truth images, while Our (\textsc{+Gt}) uses ground-truth texts). Among methods not using ground-truth data, the best accuracy is highlighted in \first{blue}, followed by the second-best in \second{black}.}
    \label{tab:text_zoom}
    \vspace{-4mm}
\end{table*}

\begin{table}[t]
    \footnotesize
    \centering
    \setlength\tabcolsep{6pt}
        \parbox{.495\linewidth}{
        \begin{tabular}{lccc}
            \toprule
                \textbf{OCR} & $\bm \omega$ & $\bm R$ & \textbf{Acc.$\uparrow$} \\
            \midrule
                \footnotesize{\textsc{Img}} & \multirow{2}{*}{0.0} & \multirow{2}{*}{-} & \multirow{2}{*}{57.8\%} \\
                \footnotesize{\textsc{Only}}&&& \\
            %%%%%%%%%%%% CRNN %%%%%%%%%%%% 
            \midrule
                \multirow{6}{*}{\footnotesize{\textsc{+Crnn}}} & \multirow{3}{*}{1.0}
                  & 0 & 39.7\% \\
                & & 1 & 57.3\% \\
                & & 2 & 57.4\% \\
            \cline{2-4}
                & \multirow{3}{*}{0.5}
                  & 0 & 52.0\% \\
                & & 1 & 57.6\% \\
                & & 2 & \first{57.8\%} \\
            \bottomrule
        \end{tabular}
    }
    \parbox{.495\linewidth}{
        \begin{tabular}{lccc}
            \toprule
                \textbf{OCR} & $\bm \omega$ & $\bm R$ & \textbf{Acc.$\uparrow$}\\
            \midrule
                \multirow{2}{*}{\footnotesize{\textsc{+Gt}}}
                & 1.0 & - & 74.6\% \\
                & 3.0 & - & \first{77.6\%} \\
            \midrule
            % %%%%%%%%%%%% MORAN %%%%%%%%%%%% 
            % \midrule
            %     \multirow{6}{*}{\textsc{+Moran}} & \multirow{3}{*}{1.0}
            %       & 0 & 44.7\% & 0.343 \\
            %     & & 1 & 58.2\% & 0.233 \\
            %     & & 2 & 58.2\% & 0.235 \\
            % \cline{2-5}
            %     & \multirow{3}{*}{0.5}
            %       & 0 & 54.2\% & 0.268 \\
            %     & & 1 & 58.6\% & 0.235 \\
            %     & & 2 & 58.8\% & 0.232 \\
            %%%%%%%%%%%% ASTER %%%%%%%%%%%% 
                \multirow{6}{*}{\footnotesize{\textsc{+Aster}}} & \multirow{3}{*}{1.0}
                  & 0 & 46.7\% \\
                & & 1 & \first{60.0\%} \\
                & & 2 & \first{60.0\%} \\
            \cline{2-4}
                & \multirow{3}{*}{0.5}
                  & 0 & 55.4\% \\
                & & 1 & 59.5\% \\
                & & 2 & 59.5\% \\
            \bottomrule
        \end{tabular}
    }
    \caption{Ablation study on the TextZoom. The CRNN and the Medium split were used for evaluation. The best accuracy in each OCR group is shown in \first{blue}.}
    \label{tab:text_zoom_ablation_study}
    \vspace{-6mm}
\end{table}

\subsection{Results on the TextZoom}
\label{sec:results:text_zoom}
% We first evaluate our model on TextZoom~\cite{textzoom}.
We finetune our TDT-pretrained multilingual model on the TextZoom~\cite{textzoom} benchmark, to compare to the STISR models focusing on Latin characters.
To ensure a fair comparison, we meticulously adhered to the training and evaluation protocols outlined in TextZoom. 
To tailor our pretrained model, we fine-tuned it for an additional 20K steps, adjusting the batch size to 128 and the learning rate to 1e-5. Furthermore, we modified the input/output image resolution from $48\!\times\!480$ to $32\!\times\!128$ and eliminated the padding strategy (for short and medium length texts),
% previously employed. This change ensures that text images with only 1-2 characters are stretched to the full 32x128 dimensions (a 1:4 ratio), 
maintaining consistency with the SOTA evaluation on TextZoom.

% OCR recognition models are used to evaluate our text restoration model. The idea is that enhanced text is easier to read for both humans and OCR models. This evaluation only needs images and their ground-truth texts.
% For performance measurement, we report the average normalized edit distance (N.E.D.):
% $\text{N.E.D.} = \frac{1}{N} \sum_{i=1}^{N} \frac{D(s_i, \hat s_i)} {\max (|s_i|, |\hat s_i|)}$,
% where $i$ indexes the example in the dataset ($N$ total examples), $s_i$ represents the Unicode text extracted from the restored text image using an OCR recognition model, $\hat s_i$ represents the ground-truth text, $D$ represents the Levenshtein Distance~\cite{levenshtein1966binary}, and $|\cdot|$ denotes the string length of the Unicode sequence.
% We further present the average OCR accuracy, determined by the proportion of $D(s_i, \hat s_i)=0$.

\vspace{1mm}
\noindent\textbf{Comparing to SOTAs.}
We included the following baseline methods: 
TSRN introduced in the original TextZoom benchmark~\cite{textzoom}; TATT \cite{Ma_2022_CVPR}, and TATT+DPMN \cite{Zhu_Zhao_Fang_Xue_2023}, which both incorporated text priors into regression models; and the latest DM-based methods published in 2023-2024: TextDiff \cite{liu2023textdiff}, TCDM \cite{noguchi2024scene}, DCDM \cite{singhdcdm}.
Alongside state-of-the-art methods, we have included Bicubic and HR baselines. These use the upscaled low-quality image and the ground-truth high-resolution image as model predictions, respectively.
The OCR evaluators employed were ASTER \cite{shi2018aster}, MORAN \cite{luo2019moran}, and CRNN \cite{shi2016end}.
Tab.~\ref{tab:text_zoom} shows the results. Our method without text conditioning, \textsc{Ours (ImgOnly}), exhibits performance (CRNN 56.6\% MORAN 63.0\% ASTER 65.4\%) comparable to the DMs (TextDiff, TCDM, DCDM). When incorporating OCR text as an additional input (\textsc{Ours (+Iter. Aster)}), performance (CRNN 58.7\% MORAN 64.0\% ASTER 65.8\%) surpasses all methods except HR and \textsc{Ours (+Gt)}. Notably, combining the low-quality image with the ground-truth reference text (\textsc{Ours (+Gt)}) yields results (CRNN 76.6\% MORAN 84.7\% ASTER 87.2\%) exceeding the HR. The qualitative results can be found in the supplementary materials.

\vspace{1mm}
\noindent\textbf{Ablation Study.}
We conducted an ablation study using the CRNN evaluator and the Medium validation split from the TextZoom dataset.
We examined the impact of various text conditions, including the scenarios with no text (\textsc{ImgOnly}), the use of reference text (\textsc{+Gt}), and the incorporation of text generated by different OCR models (\textsc{+Crnn}, \textsc{+Aster}). Additionally, we investigated the effects of the $\omega$ parameter (Sec.~\ref{sec:method:inference:cfg}) and the $R$ (Sec.~\ref{sec:method:inference:iterative}).

Tab.~\ref{tab:text_zoom_ablation_study} shows the results. First of all, the highest performance was achieved when ground-truth texts (\textsc{+Gt}) were incorporated, with accuracy exceeding 70\%. Furthermore, the high quality of the ground truth data allows for a higher emphasis ($\omega\!=\!3.0$) on the OCR-guided term in Eq.~\ref{eq:cfg}, which leads to improved restoration performance (77.6\% compared to 74.6\%).
Second, the OCR recognition model is a key factor in overall performance. For instance, with the least performant \textsc{Crnn} model (among \textsc{Aster}, \textsc{Moran}, \textsc{Crnn}), the final performance remains similar to the image-only model (57.8\%), regardless of how we adjust the other parameters ($\omega$ and $R$).
Third, the performance of STISR is significantly improved by incorporating text guidance extracted from the OCR model, which is fed with the restored image: $g^{(1)}(c_I)=g(c_I, \psi(g(c_I, \varnothing)))$. This improvement ($R=1$ vs $0$) is consistently observed across different OCR model groups, including \textsc{+Crnn} (57.3\% vs 39.7\%, 57.6\% vs 52.0\%), \textsc{+Aster} (60.0\% vs 46.7\%, 59.5\% vs 55.4\%), validating the effectiveness of the iterative inference method proposed in Sec.~\ref{sec:method:inference:iterative}. To better understand the iterative refinement, we have enclosed qualitative examples in the supplementary materials.
Finally, the value of $\omega$ in Sec. ~\ref{sec:method:inference:cfg} should be adjusted based on the quality of the OCR model (e.g., $0.5$ for \textsc{+Crnn}, $1.0$ for \textsc{+Aster}, and $3.0$ for \textsc{+Gt}). For less accurate OCR models, particularly when applied directly to the LR image to reduce computational overhead, a value of $<\!1.0$ is recommended  (at $R\!=\!0$, $\omega\!=\!0.5$ is consistently better then $1.0$). This encourages the model to rely more on the image itself, mitigating the impact of potentially inaccurate OCR predictions.

\vspace{-2mm}
\section{Conclusion}
\label{sec:conclusion}
% This paper presents TextSR, a novel approach to Scene Text Image Super-Resolution. The TextSR training process alternates between two tasks: first, reconstructing high-resolution images from their low-resolution counterparts; and second, generating high-resolution images from multilingual UTF8 characters. To address the challenges posed by inaccurate and noisy OCR during inference, we propose the use of classifier-free guidance weighting and iterative OCR conditioning. Our model has been comprehensively evaluated on both the TextZoom and TextVQA datasets. The results of our experiments demonstrate that our model significantly outperforms the current state-of-the-art methods in terms of perceived legibility of the perceptual quality. The paper also proves that language-specific design is not essential, as our single STISR model can resolve Text Super-Resolution tasks across different languages.
This paper introduces TextSR, a novel approach for Multilingual Scene Text Image Super-Resolution. TextSR represents multilingual texts using UTF-8 and steers the restoration process through cross-attention. The model learns multilingual character-to-shape diffusion priors from a large-scale text detection and transcription dataset, enabling it to generate multilingual texts even without an image. During inference, the model relies on OCR results and uses techniques to manage inaccurate and noisy OCR. Evaluation on TextZoom and TextVQA datasets shows that TextSR surpasses current methods.
{
    \small
    \bibliographystyle{ieeenat_fullname}
    \bibliography{main}
}
\clearpage

\appendix
\setcounter{page}{1}

\maketitlesupplementary

\section{Implementation Details}
\label{sec:supp:more_details}

\noindent\textbf{Affine Transformations of Text Regions.} To extract the text region $T_i$, we apply the affine transformation $\Phi(I, \theta_i)$ to the image $I$ using parameters $ \theta_i \in \mathbb{R}^{2\times3}$. For brevity, we will omit the index $i$ in this section. Given the detection box, represented by three coordinates $b^{(src)}=[(x^{(0)}, y^{(0)}), (x^{(1)}, y^{(1)}), (x^{(2)}, y^{(2)})]$ (the 4th coordinate can be inferred from the other three assuming affine transformation), and the target crop box $b^{(dst)}=[(0, 0), (0, h), (w, h)]$ (where $h$ is a constant of 48 in our paper and $w$ is set to maintain the aspect ratio), we can obtain the transformation parameter $\theta$ from $b^{(src)}$ and $b^{(dst)}$. For this purpose, we utilize the OpenCV function $\theta=$cv.getAffineTransform($b^{(src)}$, $b^{(dst)}$). Then, the pointwise transformation of the cropping can be expressed as:

\begin{equation*}
    \begin{pmatrix}
    x^t \\
    y^t \\
    \end{pmatrix} = 
    \begin{bmatrix}
    \theta_{00} & \theta_{01} & \theta_{02} \\
    \theta_{10} & \theta_{11} & \theta_{12}
    \end{bmatrix}
    \begin{pmatrix}
    x^s \\
    y^s \\
    1
    \end{pmatrix},
\end{equation*}
where $(x^t, y^t)$ is a point in the cropped image $T_i$, and $(x^s, y^s)$ is its according point in the original image $I$. The above transformation can be implemented using $\Phi(I, \theta_i)$=cv.warpAffine($I$, $\theta_i$).

The transformation to paste the restored text $T_i'$ back to the original image $I$ can be achieved using the invert affine transformation. Given 1$\times$ scaling, the following ways can be used to estimate the transformation parameter: $\theta^{-1}=$cv.getAffineTransform($b^{(dst)}$, $b^{(src)}$) or $\theta^{-1}=$cv.invertAffineTrasform($\theta$). For 2$\times$ and 4$\times$ upscaling, an additional scaling transformation is applied.

\vspace{2mm}
\noindent\textbf{Training Our Real-ESRGAN Model.} 
Following the methodology outlined in Real-ESRGAN [65], we trained two separate models for 2$\times$ and 4$\times$ upsampling.  Both models utilize the same RRDB architecture with 23 residual-in-residual dense blocks, as proposed in the original work.  However, instead of using the DF2K dataset for training, we employed the LSDIR dataset [24], which provides a richer and more diverse set of high-resolution images.  The training process was divided into two phases.  Initially, we trained the models with an L1 pixel loss and a learning rate of 0.0001 for 400,000 steps. This stage focuses on establishing a strong foundation for reconstructing high-frequency details. Subsequently, we fine-tuned the models for another 400,000 steps using a learning rate of 0.0001 and incorporating the perceptual loss components as defined in Real-ESRGAN, namely, the GAN loss with a relativistic discriminator and the perceptual loss based on VGG features. This two-stage training strategy aims to balance the preservation of low-level details with the enhancement of overall perceptual quality. This entire training pipeline was implemented using Tensorflow and A100 GPUs.

\vspace{2mm}
\noindent\textbf{Blending with Our Real-ESRGAN Outputs.} The blending algorithm, denoted by  $\oplus$  and detailed in Sec. 3, must ensure coherence between the TextSR outputs from  $g$  and the Real-ESRGAN outputs  $f(I)$. Our solution involves post-processing the TextSR output  $T_i'$. We assume $f(T_i)$ represents the results cropped from the Real-ESRGAN output $f(I)$ using the cropping parameters $\theta_i$. Therefore, a consistent prediction of the text image to replace $T_i'=g(T_i)$ in our Sec. 3 can be represented as:
\vspace{-2mm}
\begin{equation*}
    T_i' = g(T_i) - \delta(g(T_i), f(T_i)),
\vspace{-2mm}
\end{equation*}
where the $\delta (g(T_i), f(T_i))$ quantifies the discrepancies in color between $g(T_i)$ and $f(T_i)$. $\delta$ can be estimated using
\begin{equation*}
    \delta (g(T_i), f(T_i)) = \text{LPF}(g(T_i)) - \text{LPF}(f(T_i)),
\vspace{-2mm}
\end{equation*}
where ``LPF'' refers to a lowpass filtering process, which is carried out using a Gaussian filter that has a sigma value of 3.0.
Therefore, our final prediction is
\vspace{-2mm}
\begin{equation*}
    T_i' = \underbrace{\text{LPF}(f(T_i))}_{\text{from our Real-ESRGAN}} + \underbrace{\Bigl[g(T_i) - \text{LPF}(g(T_i))\Bigr]}_{\text{from TextSR}},
\vspace{-2mm}
\end{equation*}

% Our final config from cl/694211302:
% /cns/is-d/home/luma/yek/deep_upscaler/rs=6.3/models/zoom_enhance_v3/ocr_scale1_color3_cfgi1_0_cfgt3_0.pbtxt

One way to understand the equation is that the Real-ESRGAN prediction provides the low frequency elements, such as color and basic structure, while the TextSR prediction contributes the high frequency details, which enhance the overall result.

\vspace{2mm}
\noindent\textbf{Iterative OCR Conditioning.}
We elaborate the iterative OCR conditioning algorithm referenced in our main paper. This algorithm takes a low-resolution image ($c_I$) and a user-defined parameter $R$ specifying the number of iterations. It then produces a super-resolution image by leveraging both an OCR model $\psi$ and our TextSR model $g$.

% see `_iterative_restore` from https://github.com/yekeren/TextZoom/blob/text_zoom_evaluation/src/interfaces/super_resolution.py
\begin{algorithm}
\caption{$R$-iterative OCR conditioning \\
\textbf{Inputs}: {$c_I$ - the LR image, $R$ - the number of iterations} \\
\textbf{Output}: the super-resolved image}
{\fontsize{8}{8}\selectfont
\begin{algorithmic}[1]
\If {$R = 0$}
    \State $c_T \gets \psi(c_I)$ \Comment{Run OCR on LR image}
    \State \textbf{return} $g(c_I, c_T)$
\Else
    \State $c_I' \gets g(c_I, \varnothing)$ \Comment {Restore using the image-only model}
    \For {$i \gets 1$ \textbf{to} $R$}
        \State $c_T \gets \psi(c_I')$ \Comment{Run OCR on restored image}
        \State $c_I' \gets g(c_I, c_T)$ \Comment{Restore using LR image and updated text}
    \EndFor
    \State \textbf{return} $c_I'$
\EndIf
\end{algorithmic}
}
\end{algorithm}

\section{STISR with Multilingual Guidance}
\begin{figure}[ht!]
    \centering
    \includegraphics[width=1.0\linewidth]{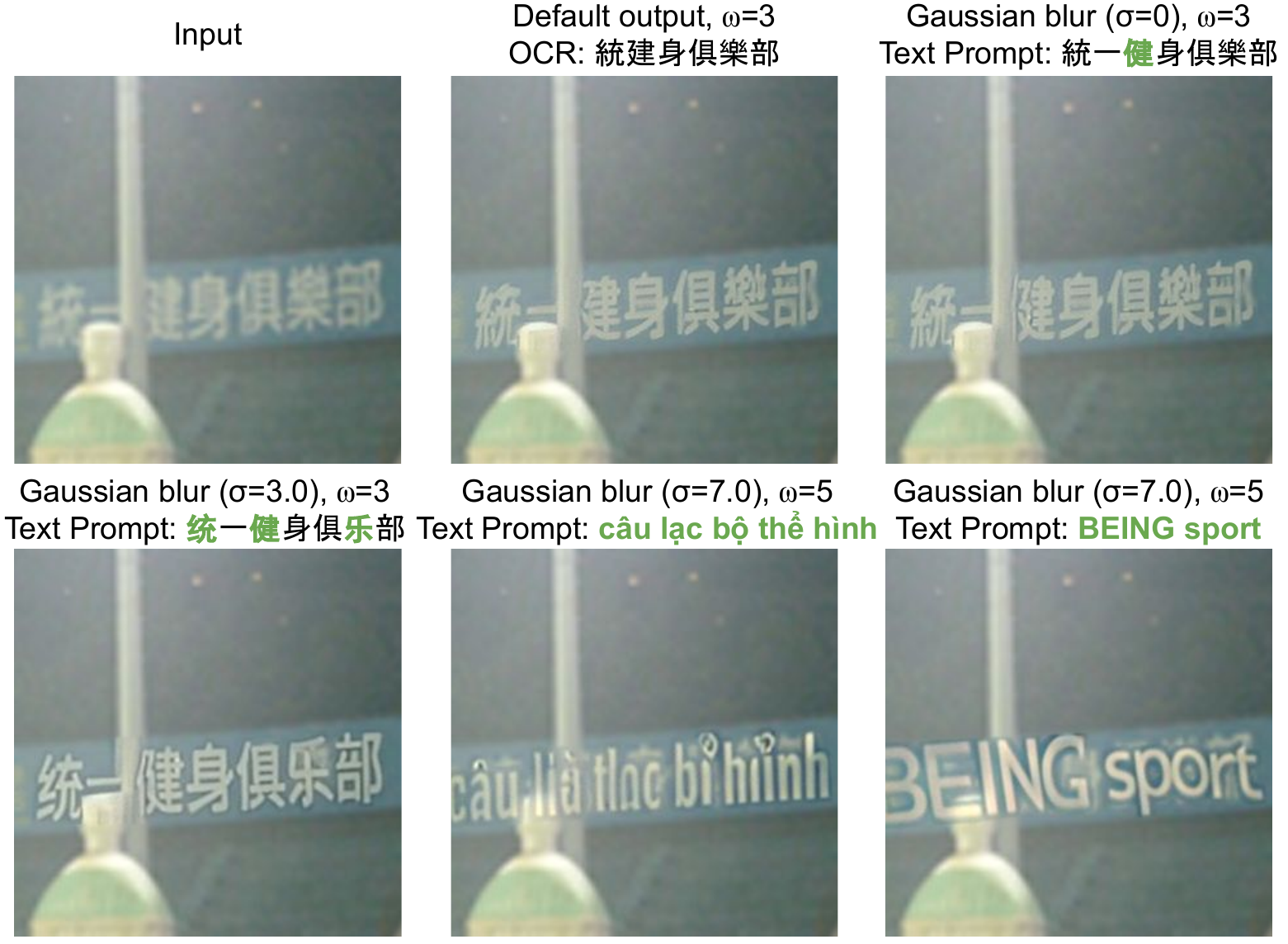}
    \caption{Prompting our model with multilingual texts. Characters that influence the image region are highlighted in \textcolor{olive}{\textbf{green}}. Our model is trained on a diverse range of languages and scripts, including Traditional Chinese, Simplified Chinese, Vietnamese, Latin characters, and so on. This allows our model to leverage the multilingual conditions to guide the restoration process.}
    \label{fig:text_editing}
\end{figure}

We conducted further qualitative study to understand how multilingual text guidance influences our model. To reduce reliance on the low-quality image, we took two steps. First, we increased the weight of the combined image-text conditioning by adjusting the parameter $\omega$ to values of 3 and 5. This emphasizes the importance of using both image and text data together. Second, we introduced more ambiguity into the visual input by applying a Gaussian blur with $\sigma$ values of 3 and 7 to the input image. This forced the model to rely less on the LR image details. Fig.~\ref{fig:text_editing} shows the results. Our findings indicate that the model effectively leverages text prompts to restore unclear text within an image.
% This indicates that the model has effectively utilized the character-to-shape diffusion priors. In cases where visual information is ambiguous, our model leverages external information. This process could be further enhanced by utilizing a larger context. For instance, OCR technology can employ graph neural networks to facilitate information sharing across different text regions~\cite{khanfir2024graph,carbonell2021named,riba2019table}.

\section{Qualitative Results on the TextVQA}
\label{sec:supp:text_vqa}

\begin{figure*}[t]
    \centering
    \includegraphics[width=1.0\linewidth]{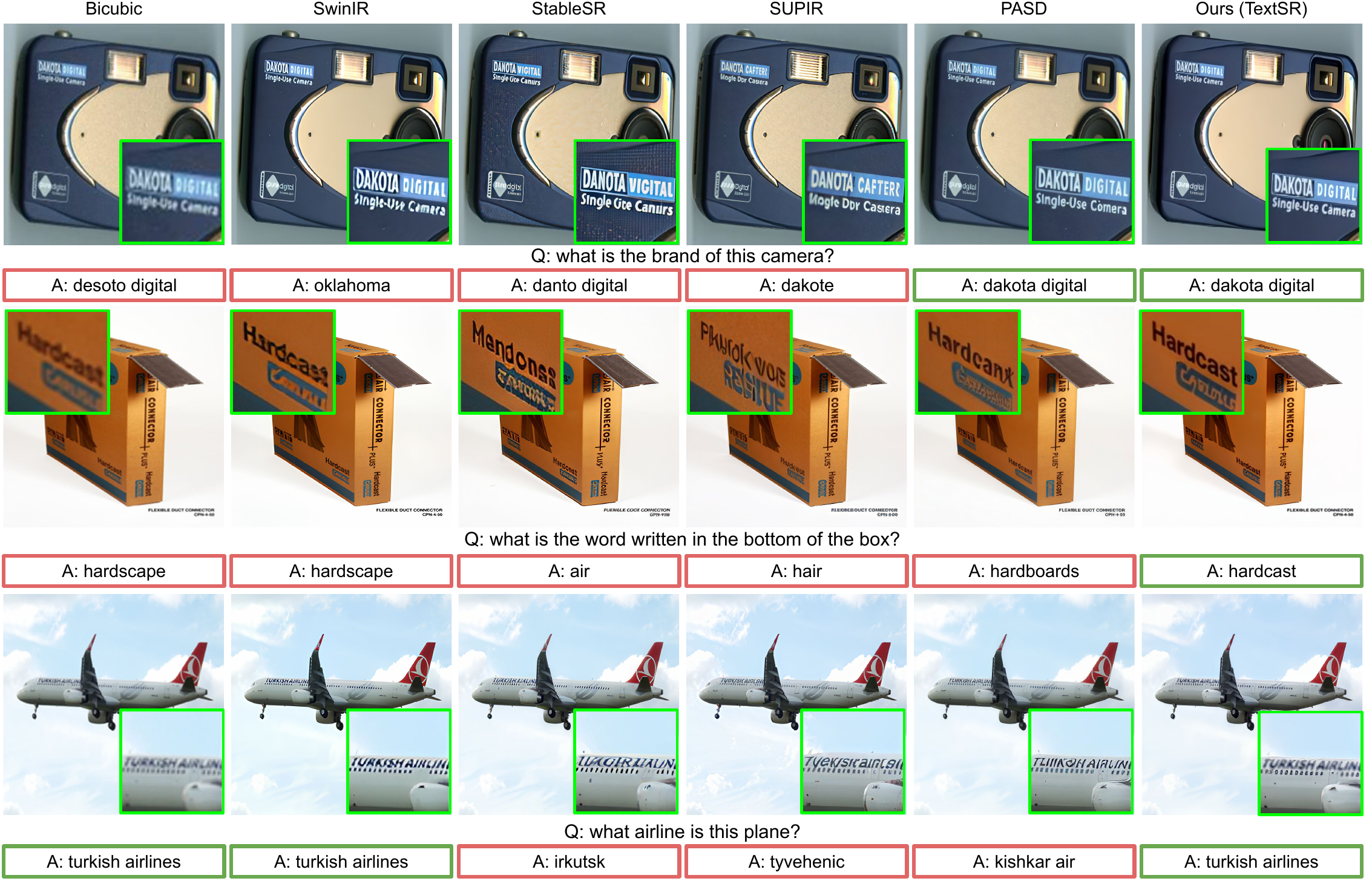}
    \caption{Answers generated by various models on the TextVQA}
    \label{fig:supp:text_vqa_qa_examples}
\end{figure*}

\begin{figure*}[t]
    \centering
    \includegraphics[width=1\linewidth]{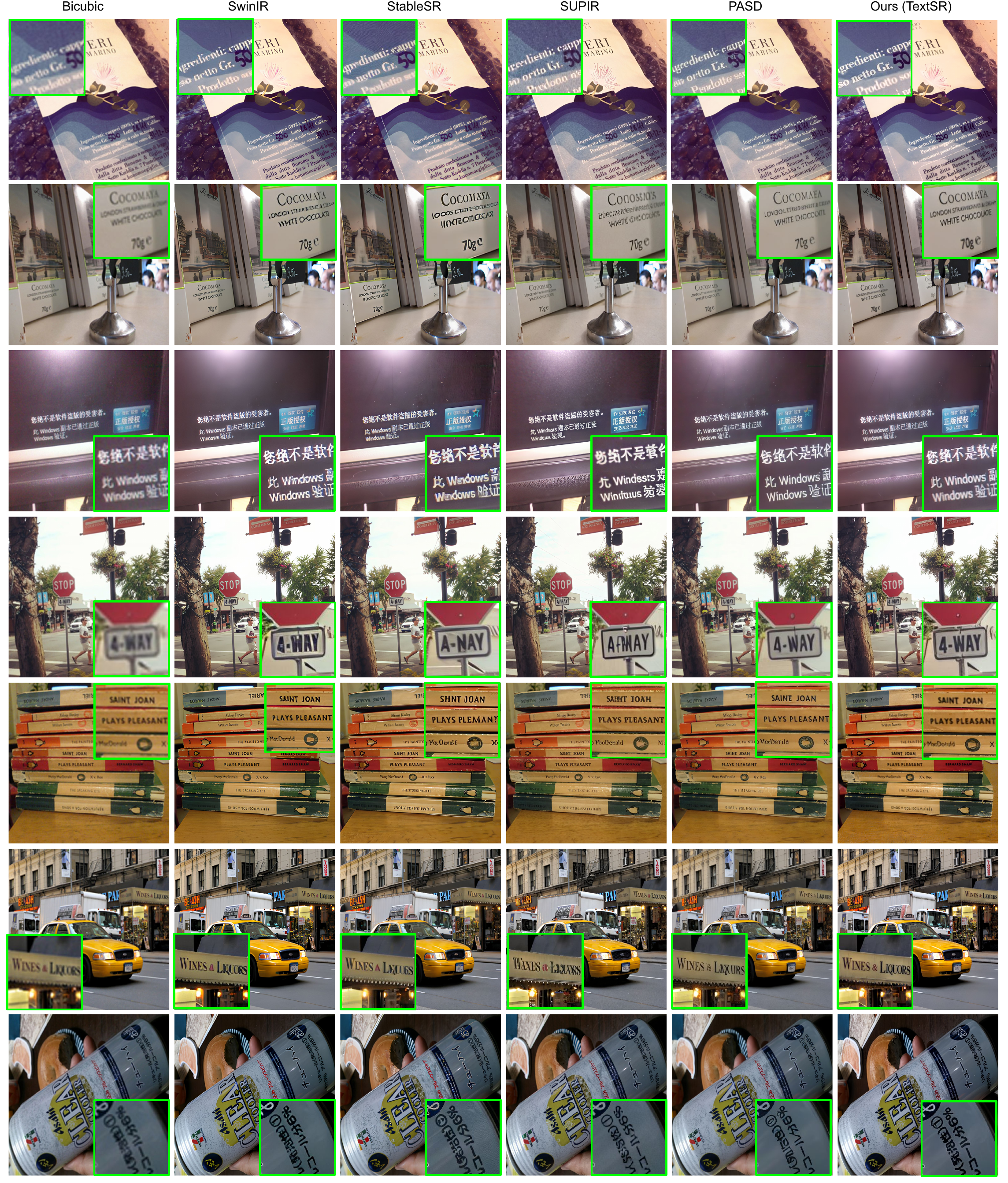}
    \caption{Qualitative results on the TextVQA.}
    \label{fig:supp:text_vqa}
\end{figure*}

Fig.~\ref{fig:supp:text_vqa_qa_examples} shows why poor restoration results cause the VQA model GIT [51]to fail.
% https://luma-scope.googleplex.com/529280003
Fig.~\ref{fig:supp:text_vqa} presents more qualitative results on the TextVQA.

\begin{figure*}[t]
    \centering
    \includegraphics[width=1\linewidth]{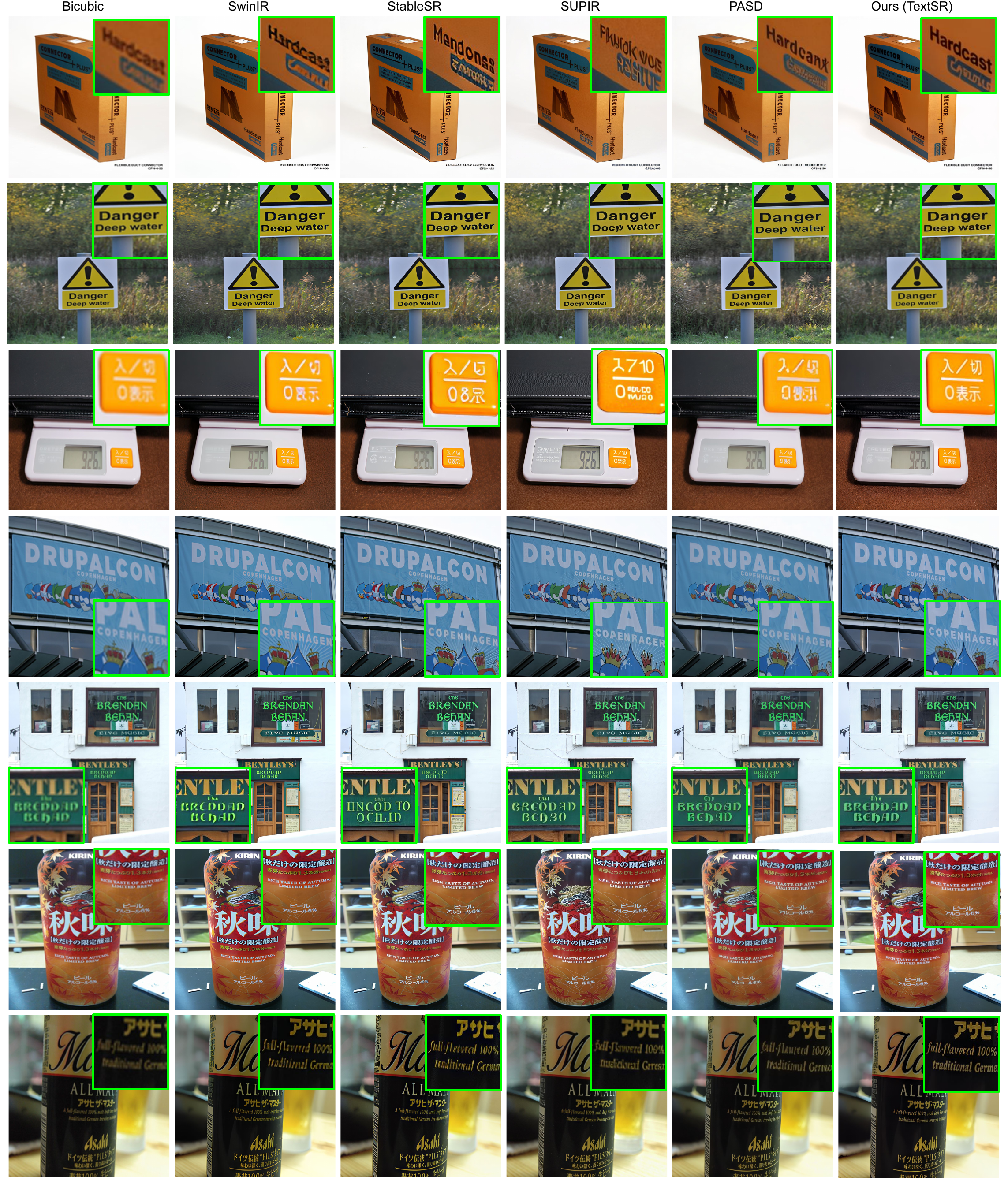}
\end{figure*}

\section{Qualitative Results on the TextZoom}
\label{sec:supp:text_zoom}

\begin{figure*}[t]
    \centering
    \includegraphics[width=1\linewidth]{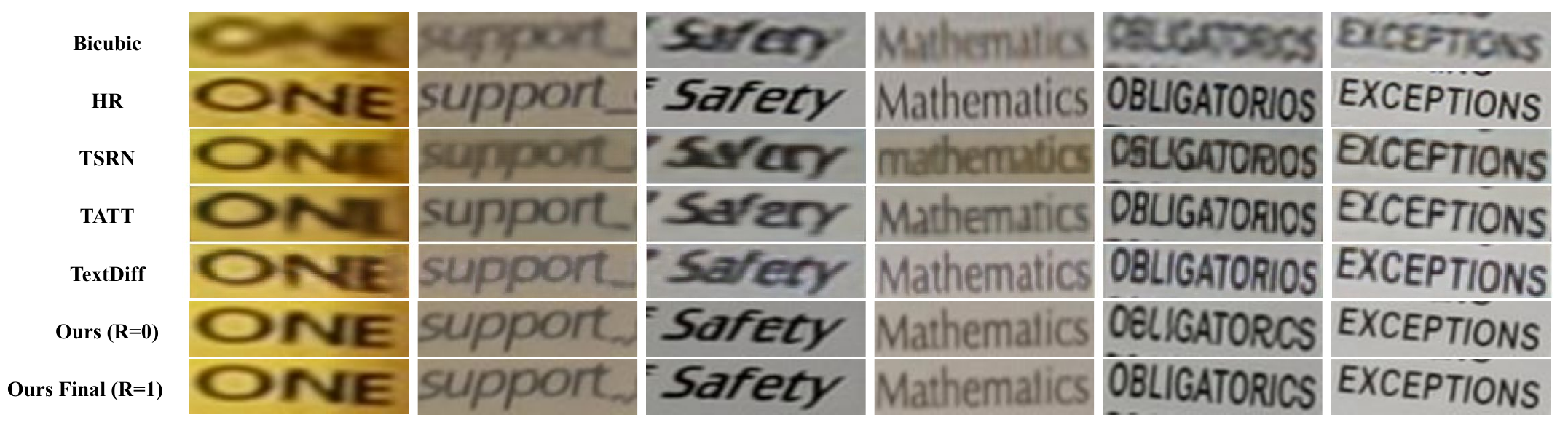}
    \caption{Comparing to the SOTAs on the TextZoom.}
    \label{fig:supp:text_zoom_vs_sotas}
\end{figure*}

\begin{figure*}[t]
    \centering
    \includegraphics[width=1\linewidth]{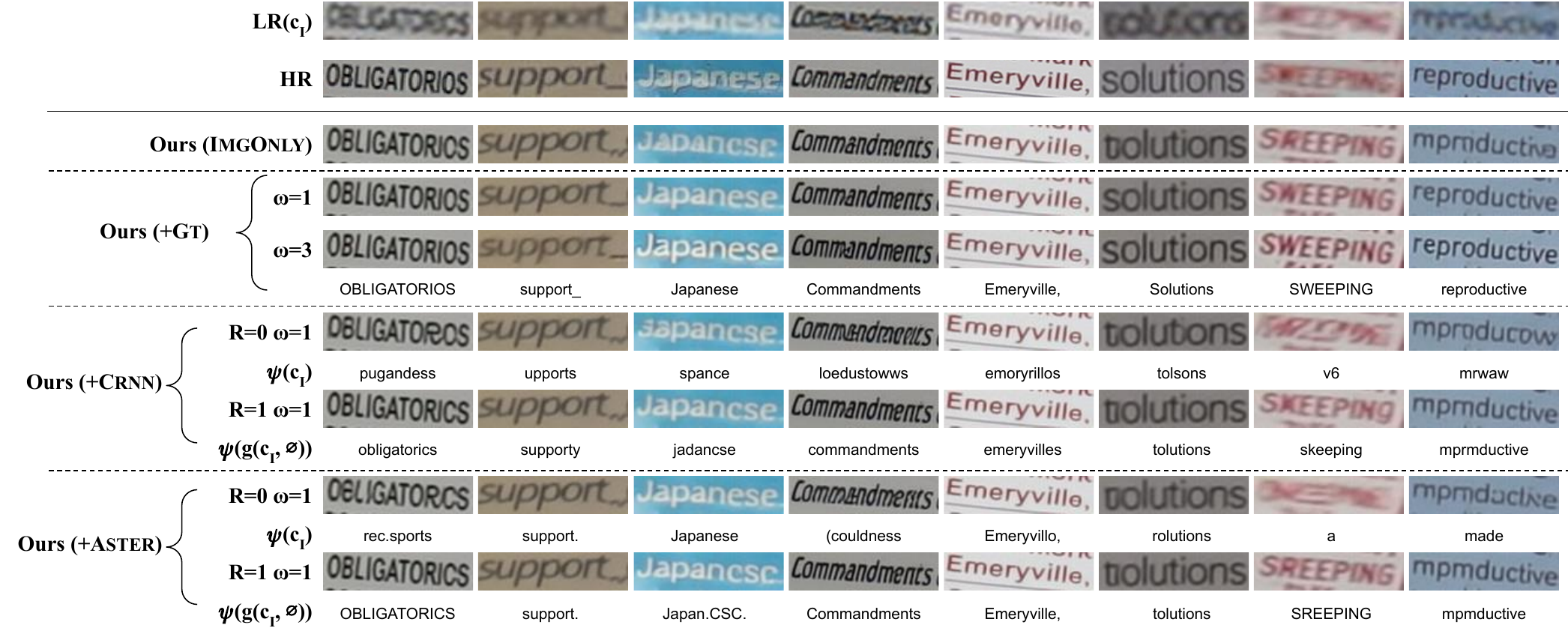}
    \caption{Qualitative results on the TextZoom. The text displayed beneath the images represents the final text prompt provided to our TextSR model, following the application of iterative OCR conditioning.}
    \label{fig:supp:text_zoom}
\end{figure*}

In Fig.\ref{fig:supp:text_zoom_vs_sotas}, we show the comparisons to the SOTAs.
The effect of the proposed Classifier-Free Guidance control and the Iterative OCR conditioning on the TextZoom images are presented in Fig.~\ref{fig:supp:text_zoom}.

% \clearpage

% \section{Image attribution}
% This paper includes several images from the TextVQA dataset. Tab.~\ref{tab:supp:license} contains the Flickr links and licensing information for these images. We extend our gratitude to the photographers who generously shared their work.

% \input{tabs/license}

\end{document}